\documentclass[lettersize,journal]{IEEEtran}
\usepackage{amsmath,amsfonts}
\usepackage{algorithmic}
\usepackage{array}
\usepackage{booktabs}
\usepackage[caption=false,font=normalsize,labelfont=sf,textfont=sf]{subfig}
\usepackage{textcomp}
\usepackage{stfloats}
\usepackage{url}
\usepackage{verbatim}
\usepackage{graphicx}
\usepackage{float} 
\usepackage{multirow}
\usepackage{makecell}
\def\BibTeX{{\rm B\kern-.05em{\sc i\kern-.025em b}\kern-.08em
    T\kern-.1667em\lower.7ex\hbox{E}\kern-.125emX}}
\usepackage{balance}
\begin{document}
\title{DexLink Hand: A Compact, Affordable, 16-DOF Linkage-Driven Hand with Human-Like Dexterity}

\author{Hao Wu$^{1,2}$, Yanzhe Wang$^{1}$, Yu Feng$^{2}$, Jian Liu$^{2}$, Jihao Li$^{1}$, \\Jianshu Zhou$^{2}$,~\IEEEmembership{Member,~IEEE,~ASME}, and Huixu Dong$^{1}$,~\IEEEmembership{Member,~IEEE,~ASME}

\thanks{Hao Wu and Yanzhe Wang contribute to this work equally. 1.\,Department of Mechanical Engineering, Zhejiang University, Hangzhou, China. 2.\,Department of Mechanical Engineering, National University of Singapore, Singapore. Corresponding author: Huixu Dong (huixudong@zju.edu.cn)
}}

\markboth{}
{DexLink Hand: A Compact, Affordable, 16-DOF Linkage-Driven Hand with Human-Like Dexterity}

\maketitle

\begin{abstract}

Dexterous robotic hands face a longstanding trade-off among dexterity, compactness, and affordability. Particularly, high-degree-of-freedom designs typically demand complex actuation and transmission, hindering integration into human-scale forms. To address these challenges, this work presents a compact, low-cost linkage-driven anthropomorphic hand that achieves high dexterity, structural integration, and human-hand-like functionality. The hand integrates 20 joints driven by 16 independent actuators, with all actuation, sensing, and transmission components compactly embedded within a human-hand-sized structure. The resulting prototype weighs only 320g at a total cost below USD 400. To meet these objectives, a hybrid mechanical architecture combining planar and spatial linkage mechanisms is proposed, enabling decoupled multidirectional motion, biomimetic joint synergies, and high passive load-bearing capability. The thumb further incorporates biomimetic features supporting human-like reconfiguration and opposition movements. Through the coordinated integration of these mechanisms and structural layout, the prototype achieves a highly integrated design with anthropomorphic dexterity. Experimental evaluations demonstrate that the hand achieves the maximum Kapandji score, reproduces all 33 Feix grasp types, and performs stable grasping and dexterous manipulation across a wide variety of daily objects and tools. These results validate the proposed hand as an affordable, compact, and mechanically efficient platform for dexterous manipulation, teleoperation, and robot learning in human-centered environments.

\end{abstract}

\def\abstractname{Note to Practitioners}
\begin{abstract}
Deploying dexterous robotic hands in real-world applications remains challenging because enhancing dexterity typically requires additional degrees of freedom, which in turn increases hardware cost, system complexity, and physical size. This work presents a design methodology that leverages optimized linkage-based mechanisms to embed motion coordination, decoupling, and load-bearing characteristics directly into the mechanical structure of the anthropomorphic hand. By combining coupled-joint linkages, compact spatial spherical linkage mechanisms, and self-locking transmissions, the proposed approach achieves mechanically efficient motion transmission and enables a 16-degree-of-freedom hand to deliver human-like dexterity within a human-hand-scale platform. The design further promotes anthropomorphism in terms of appearance, structural layout, kinematic characteristics, and functional capabilities, thereby facilitating the development of robotic hands that can better interact with human-centric objects, tools, and environments. The resulting design principles provide a practical reference for developing dexterous manipulators that balance dexterity, payload capacity, compactness, and affordability. The proposed mechanism design and system architecture are currently being adopted in the development of a commercial dexterous hand platform. Future work will focus on integrating tactile sensing and teleoperation capabilities, further evolving the system into a sensorized platform for dexterous manipulation and robot learning.
\end{abstract}


\section{Introduction}

\begin{table*}[t]
\centering
\caption{Quantitative comparison of representative dexterous robotic hands.}
\label{tab:comparison}

\renewcommand{\arraystretch}{1.25}
\setlength{\tabcolsep}{3pt}
\footnotesize

\resizebox{\textwidth}{!}{%
\begin{tabular}{p{2.5cm} c p{2.2cm} c p{1.8cm} c c p{3.8cm}}
\hline
\textbf{Platform}\rule{0pt}{5.0ex} &
\textbf{\shortstack{Actuated\\DOFs}} &
\textbf{Size (mm)} &
\textbf{Weight} &
\textbf{\shortstack{Approx. Cost}} &
\textbf{\shortstack{Fingertip\\Force}} &
\textbf{\shortstack{Payload}} &
\textbf{\shortstack{Transmission}} \\
\hline

DLR/HIT Hand II \cite{liu2008multisensory}
& 15
& Larger than human
& $\sim$1.5\,kg
& Very high
& 10\,N
& N/A
& Tendon + differential bevel gears \\

Shadow Hand \cite{sharma2014shadow}
& 20
& 448\,mm length
& $\sim$4.3\,kg
& $>$USD 50k
& $\sim$10\,N
& $\leq$5\,kg active
& Tendon-driven \\

Allegro Hand
& 16
& $284 \times 160 \times 54$
& $\sim$1.2\,kg
& $\sim$USD 15k
& $\sim$5\,N
& $\leq$5\,kg active
& Direct geared DC motors \\

ILDA Hand \cite{kim2021integrated}
& 15
& $218 \times 90 \times 56$
& 1.1\,kg
& High
& 34\,N
& 18\,kg
& Linkage-driven \\

RBO Hand 3 \cite{puhlmann2022rbo}
& 16
& Human scale
& $\sim$1.6\,kg
& High
& $\sim$8.3\,N
& $\sim$3.9\,kg passive
& Pneumatic soft actuation \\

Leap Hand \cite{shaw2023leap}
& 16
& $\sim 300 \times 100$
& $\sim$350\,g
& $\sim$USD 2k
& $\sim$5\,N
& N/A
& Direct geared servos \\

DexHand 021 \cite{yuan2025development}
& 12
& $296 \times 113 \times 56$
& $\sim$1\,kg
& $\sim$USD 16k
& $\sim$12\,N
& 5\,kg active
& Tendon-driven rope-elastic \\

ORCA Hand \cite{christoph2025orca}
& 17
& Human scale
& $\sim$1.3\,kg
& $<$CHF 2k
& N/A
& $\sim$10.5\,kg passive
& Tendon-driven \\

RUKA Hand \cite{zorin2025ruka}
& 11
& $\sim$180\,mm length
& $\sim$500\,g
& $\sim$USD 1.3k
& 2.74\,N
& $\sim$6\,kg passive
& Tendon, underactuated \\

Krysalis Hand \cite{basheer2025krysalis}
& 18
& $240 \times 92$
& 790\,g
& High
& 10\,N
& $>$4.5\,kg passive
& Integrated geared actuation \\

CYJ Hand \cite{chai2025customize}
& 22
& Forearm-integrated
& $\sim$750\,g
& $<$USD 200
& Low
& $\sim$10\,kg passive
& Tendon-driven \\

Unitree Dex5-1
& 16
& $217 \times 128 \times 72$
& $\sim$1.1\,kg
& $\sim$USD 4--6k
& $\sim$10\,N
& 3.5--4.5\,kg active
& Geared motor-driven joints \\


CasiaHand \cite{yan2025casiahand}
& 7
& $245 \times 93 \times 56$
& $\sim$890\,g
& Low
& 16.43\,N
& N/A
& Tendon, underactuated \\

ISyHand \cite{richardson2025isyhand}
& 18
& $255 \times 130 \times 38$
& $\sim$620\,g
& $\sim$USD 1.3k
& $\sim$5.5\,N
& $\sim$9\,kg passive
& On-joint servo-driven \\

\hline
\textbf{This Work}
& \textbf{16}
& $\mathbf{190 \times 88 \times 55}$
& \textbf{320\,g}
& \textbf{$<$USD 400}
& \textbf{$>$6\,N}
& \textbf{$>$10\,kg passive}
& \textbf{Linkage-driven + worm gear} \\
\hline

\end{tabular}%
}
\end{table*}

\IEEEPARstart{T}{he} human hand is widely recognized as one of the most complex parts of the human body, serving as a critical medium for object manipulation and environmental interaction. Its high degree of freedom and precise muscle-joint coordination endow it with remarkable flexibility and adaptability in complex grasping and operation tasks. Motivated by this, dexterous robotic hands have long been pursued to replicate this generalizable manipulation capability across different scenarios. From multi-fingered grippers \cite{zhou2024dexterous,guo2025enabling,li2024under,wang2025flexible} to anthropomorphic hands \cite{liu2008multisensory,kim2021integrated,sharma2014shadow,puhlmann2022rbo}, a prevailing design trend is to increase the number of degrees of freedom (DOFs) to approximate the kinematic capabilities and functionality of the human hand. Representative systems, such as the DLR/HIT Hand II \cite{liu2008multisensory} and the Shadow Dexterous Hand \cite{sharma2014shadow}, employ fully actuated architectures combined with dense sensing and transmission systems to achieve high dexterity and precise control. Specifically, the Shadow Hand incorporates 20 actuated DOFs, enabling it to replicate a wide range of human-like gestures and perform in-hand manipulation tasks such as rolling a ball or reorienting a pen that are infeasible for low-DOF grippers. Similarly, the Leap Hand \cite{shaw2023leap} features a 16-DOF fully actuated, compact, and cost-efficient design, improving accessibility for large-scale data-driven learning. In general, higher DOFs tend to expand the kinematic workspace and dexterous capability of robotic hands, and have thus become a predominant approach to narrowing the functional gap with the human counterpart.

\begin{figure}[!t]
    \centering
    \includegraphics[width=1\linewidth]{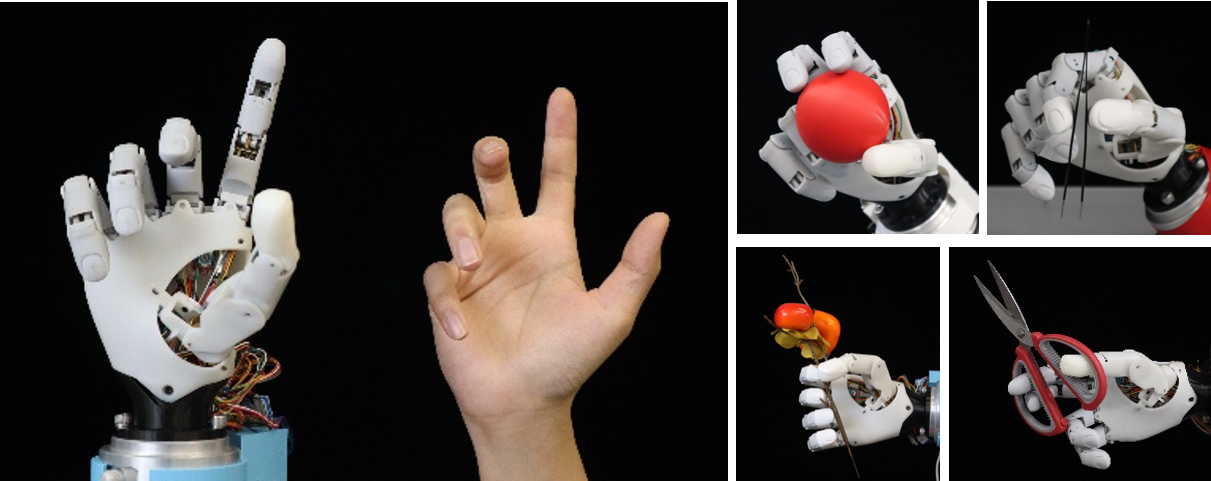}
    \caption{The prototype of the dexterous hand with an appearance and functionality akin to that of the human hand.}
    \label{FigureLabel1}
\end{figure}

However, this enhancement comes at significant costs in terms of physical size, costs, and structural complexity, resulting in limited applications in real-world scenarios. Each additional DOF typically requires an independent actuator, transmission mechanism, and sensory feedback system, which collectively increase the hand's overall size and introduce greater complexity to its structural integrity. For example, the CYJ Hand \cite{chai2025customize}, despite achieving 22 DOFs with a structural hardware cost under USD 60 (excluding actuators), still requires a forearm-integrated actuation system weighing 750g to realize its claimed dexterity, with the actuator and control system costs far exceeding the structural cost. The DexHand 021 \cite{yuan2025development}, a representative 19-DOF commercial prototype, weighs 1kg and measures 296.2mm $\times$ 113.2mm $\times$ 56.5mm, far exceeding the dimensions of the human hand, while its commercial price reaches USD 16000, primarily due to the expensive actuation modules and multi-modal sensory systems. Efforts to reduce cost through open-source and 3D-printable designs also reveal unavoidable trade-offs. Although the ORCA Hand \cite{christoph2025orca} and the RUKA Hand \cite{zorin2025ruka} achieve low cost and human‑comparable form factors, their reliance on tendon‑driven actuation results in a bulky forearm structure that houses all actuation modules and sacrifices direct joint‑angle perception. In contrast, the Krysalis Hand \cite{basheer2025krysalis} adopts a more integrated actuation to reduce transmission uncertainties; however, this also greatly increases structural complexity and overall dimensions (240.5mm $\times$ 92mm). These examples demonstrate that dexterous hands still struggle to balance high dexterity with compact structure and low cost.

An alternative strategy for reducing the size, cost, and mechanical complexity of dexterous robotic hands is to employ underactuated or synergy-driven architectures, in which coordinated multi-joint or multi-finger motions are generated through a reduced number of actuators \cite{xiong2016design,sun2021design}. By exploiting biomechanical synergies and passive mechanical coordination, these approaches can effectively reproduce a broad range of common grasping behaviors while significantly simplifying the actuation and transmission systems. Nevertheless, the reduced actuation dimensionality inherently limits independent joint controllability and motion reconfigurability, thereby constraining the dexterous versatility required for complex in-hand manipulation, generalized object interaction, and task-oriented tool-use operations.

From a mechanical perspective, this trade-off is closely related to the design of joint transmission mechanisms, especially in achieving coordinated and multi-DOF motion within compact spatial constraints. In the realm of coupled joint motion, Yoon \cite{yoon2017underactuated} presented an underactuated finger mechanism that utilizes contractible slider-cranks and stackable four-bar linkages to achieve adaptive and efficient grasping with reduced actuation requirements. Xu \cite{xu2013design} proposed a mechanical design featuring flexible shafts connected to worm gears and spur gear trains to drive the finger joints. This arrangement enables joint coupling, ensuring coordinated motion between paired joints and synchronized abduction of multiple fingers, thus achieving desired postural synergies. In contrast, the LISA Hand \cite{jin2012lisa} achieves underactuation through a linkage-based indirect self-adaptive mechanism, harnessing inverse forces from grasped objects to drive subsequent joint actions without additional motors. Meanwhile, the realization of 2-DOF in the finger base joints is crucial for enhancing manipulation capabilities. The DLR/HIT II Hand \cite{liu2008multisensory} and Rapid Hand \cite{wan2025rapid} both employ a differential gear mechanism, where two actuators share the load and collectively generate 2-DOF motion through synchronized and counter-rotational movements of bevel gears. Similarly, the ILDA Hand \cite{kim2021integrated} achieves dexterous control of 2-DOF movements through a hybrid serial-parallel linkage mechanism driven by two linear motions. Although these structures enable independent movement in two directions, they typically require the simultaneous operation of two motors, inherently increasing control complexity and limiting the flexibility in DOF configuration. To address this limitation, Li \cite{li2023linkage} proposed a differential linkage mechanism enabling effective decoupled actuation, though its large spatial footprint restricts its practicality in space-constrained applications.

Motivated by the unresolved issues in the aforementioned research areas, this study proposes a novel anthropomorphic dexterous hand that achieves human-like functionality while maintaining structural compactness, low cost, and high dexterity. The prototype achieves several competing requirements, including simplicity, lightweight construction, high DoFs, large range of motion (RoM), structural compactness, anthropomorphism, high payload capacity, and affordability. With this high dexterity and human-hand scale, the hand can handle everyday objects with a rich variety of grasp patterns, and effectively adapt to and operate common tools. Specifically, to materialize these design goals, we highlight the following contributions: (1) A crossed four-bar linkage mechanism that replicates the natural coupled motion of the human finger's proximal interphalangeal (PIP) and distal interphalangeal (DIP) joints with minimal actuation complexity. (2) A compact 2-DOF metacarpophalangeal (MCP) joint design that effectively decouples flexion/extension (Flex/Ext) and abduction/adduction (Abd/Add) movements through the coordinated use of spatial and planar linkage systems. This configuration achieves independent control of each DOF while maintaining structural compactness within the confined space of the palm. (3) A biomimetic thumb structure that preserves the functionality of the human thumb, particularly the reconfigurable relationship between the carpometacarpal (CMC) joint and the palm, enabling versatile opposition and adaptive grasping. (4) An overall biomimetic design, DOF configuration, and structural layout that closely emulate the appearance and functionality of the human hand, achieving a high degree of anthropomorphism while preserving mechanical simplicity and efficiency. Furthermore, the electronics and control system are implemented to achieve closed-loop position control of each joint. By embedding desired human-like kinematic behaviors directly into the mechanism design, the proposed hand achieves a highly optimized and fully integrated architecture that combines structure, actuation, sensing, and control within a lightweight, human-hand-scale platform. Experimental results validate the prototype's dexterity, high payload performance, and practical grasping and manipulation capabilities. All these features establish it as an affordable and reliable platform for robot learning and teleoperation.

\section{Mechanism Design}

\subsection{Design Objectives}

The primary design objective of this dexterous hand is to achieve human-like functionality within a human-hand-sized form factor, while maintaining structural simplicity, low cost, and high dexterity. Three guiding principles underpin the overall mechanism design.

First, the design emphasizes structural compactness and high integration. A primary objective is to overcome the large size and bulky architecture commonly observed in high-DOF dexterous hands by integrating miniature actuators and efficient transmission mechanisms within a human-hand-scale structure while preserving high-degree-of-freedom dexterity. Second, the hand is designed to replicate the key kinematic and morphological characteristics of the human hand. This includes the overall anthropomorphic appearance, joint distribution, thumb opposition capability, coordinated finger motion, and human-like motion trajectory, enabling natural grasping behaviors and versatile manipulation performance. Third, the system prioritizes lightweight construction, low cost, and ease of reproduction. By employing simplified linkage-driven mechanisms and accessible manufacturing methods, the hand is intended as an affordable and practical platform for teleoperation, robot learning, and research in dexterous manipulation. Based on these principles, the following sections describe the detailed mechanical implementation of the hand.

\begin{table}[t]
\centering
\caption{Average Cost Per Hand}
\label{tab:Cost}
\begin{tabular}{lccc}
\toprule
\textbf{Item} & \textbf{Unit Cost} & \textbf{Number} &\textbf{Total}\\ 
\midrule
$\phi$12 motor  & 5  & 10 & 50\\
$\phi$6 motor   & 15  &  1 & 15\\
KST servo motor  & 35  & 4 & 140\\
Worm gear & 1.5  & 7 & 10.5\\
Bearing & 0.5  & 5 & 2.5\\
3D-printed parts & 50  &  \textbackslash & 50\\
CNC machined parts & 80 & \textbackslash & 80 \\
Dowel pins and screws & 6 &  \textbackslash & 6\\
Hall-effect encoder & 5  & 2 & 10\\
PCB boards & 5  & 2 & 10\\
ESP32 controller & 3  & 2 & 6\\
Electronic wire & 1  & 10 & 10\\
\midrule
Total Cost &  &  & 390\\
\bottomrule
\end{tabular}
\end{table}

\subsection{Hand Mechanism}

\begin{figure*}[!htbp]
    \centering
    \includegraphics[width=1\linewidth]{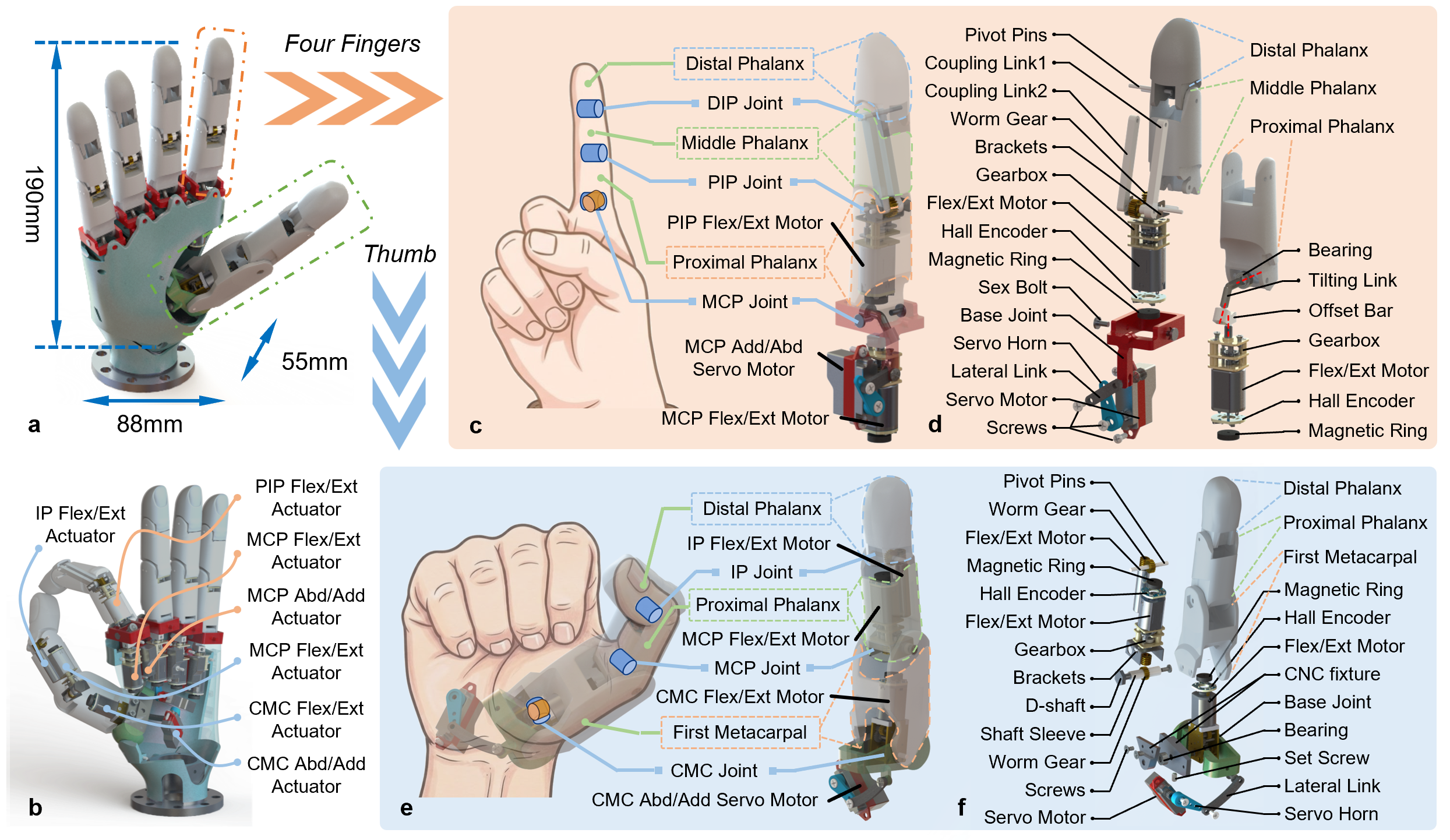}
    \caption{Demonstration of the dexterous hand's structure and mechanisms. (a) The overall structure of the hand, with dimensions of approximately 190mm x 88mm x 55mm and a finger width of 17mm, closely resembling the size of an adult human hand; (b) Mapping between joints and actuators; (c) Configuration of the four fingers; (d) Exploded view of the finger structure; (e) Configuration of the thumb; (f) Exploded view of the thumb structure.}
    \label{FigureLabel2}
\end{figure*}

The dexterous hand consists of a palm and five articulated fingers, with 20 joints driven by 16 independent actuators. Fig. \ref{FigureLabel2}a illustrates the overall design and structural features of the hand. The hand measures approximately 190mm (length) × 88mm (width) × 55mm (depth) and weighs 320g, achieving dimensions and overall morphology comparable to those of an adult human hand. This compact, lightweight construction is realized through the use of miniature DC brushed motors for actuation and 3D-printed structural components, combined with optimized mechanical design. As a result, each finger can deliver a force output exceeding 6N and withstand passive loads of over 60N, while the whole hand can sustain a maximum payload exceeding 100N. Table I summarizes a quantitative comparison of the proposed hand with representative dexterous robotic hands.

Linkage mechanisms are extensively employed in this design to achieve precise motion control and generate complex, human‑like movement trajectories. These mechanisms enable desired coupled or decoupled motion while maintaining structural compactness. The actuation scheme and joint distribution are shown in 
Fig. \ref{FigureLabel2}b. Specifically, the thumb is driven by three Flex/Ext motors and one servo for lateral movement, while each of the remaining fingers is actuated by two Flex/Ext motors and one lateral movement servo motor. Each brushed motor is equipped with a Hall sensor for real-time angular position feedback, thus enabling closed-loop position control of the joints. ESP32 microcontrollers are employed to achieve cost-effective, synchronized closed-loop position control across all joints simultaneously, while communicating with the host computer via serial port. The actuation, transmission, and sensing subsystems are all integrated within the hand, with the control unit embedded in the wrist module, contributing to its compactness, efficiency, and structural flexibility. Moreover, this approach provides a design solution for low-cost, high-degree-of-freedom dexterous hands, with a total cost of less than USD 400. The detailed component composition of the hand and corresponding costs are provided in Table II. 

\subsection{Four-finger Mechanism}

This design leverages the anatomical similarity of the four fingers (index, middle, ring, and pinky) to streamline both the manufacturing process and associated costs by implementing an identical structural design for each. Fig. \ref{FigureLabel2}c and \ref{FigureLabel2}d illustrate the finger structure and its exploded view. The fingers are strategically positioned at varying heights on the palm to replicate the natural alignment of the human hand. Here, we highlight two typical motion characteristics of the human finger: the coupled flexion of PIP and DIP joints, and the two-degree-of-freedom motion of the MCP joint. 

\begin{figure*}[!htbp]
    \centering
    \includegraphics[width=1\linewidth]{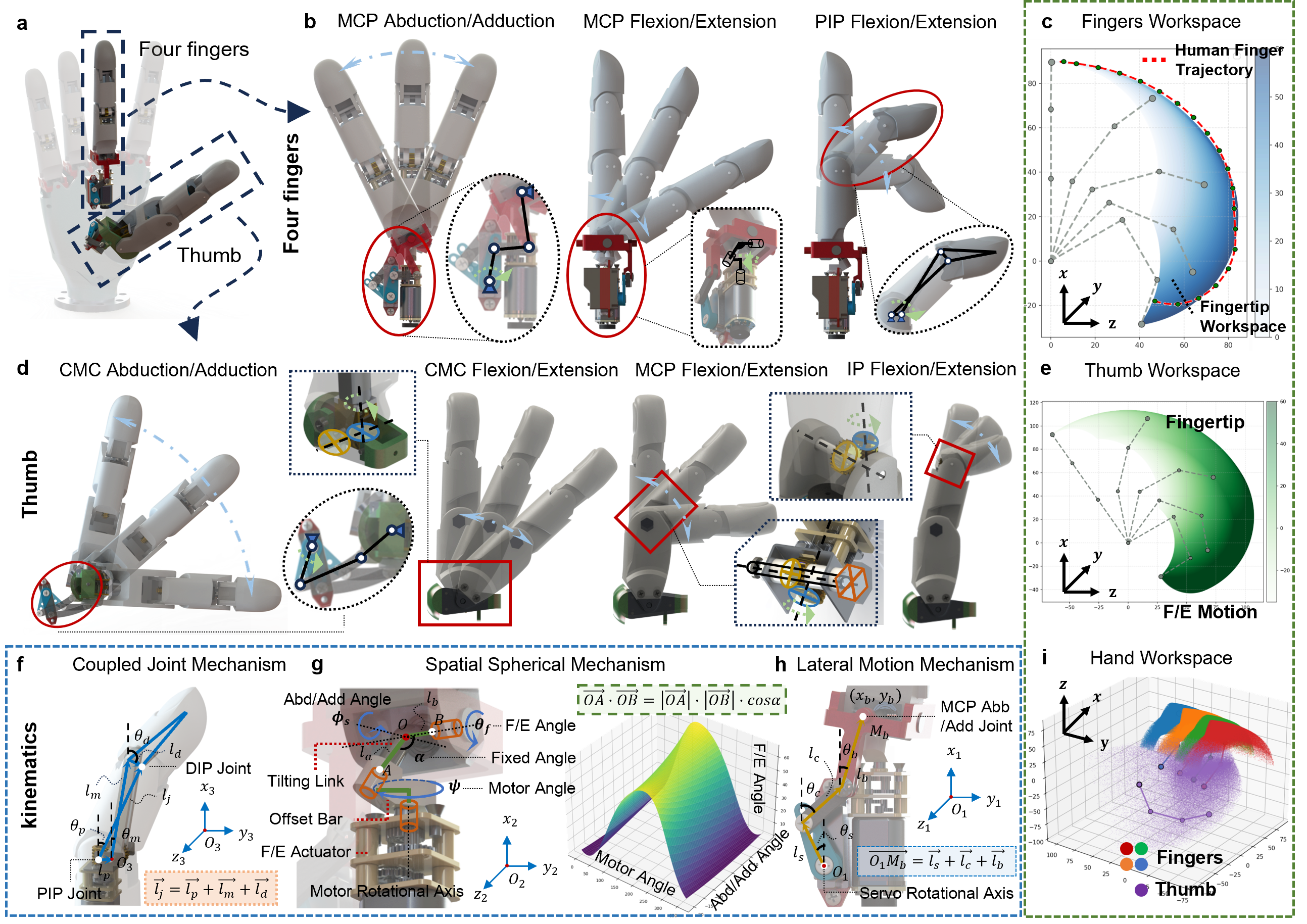}
    \caption{Transmission structure and kinematic model of the anthropomorphic hand. (a) Structure of the robotic finger designed based on the proposed mechanism. (b) Transmission mechanisms of the four fingers. (c) Workspace of the four fingertips. The red dashed curve represents the human fingertip workspace measured using a data glove. (d) Transmission mechanisms of the thumb. (e) Workspace of the thumb fingertip. (f-h) The schematic diagram of three representative linkage transmission mechanisms. (i) Workspace of the hand.}
    \label{FigureLabel3}
\end{figure*}

Firstly, the PIP and DIP joints of the human hand inherently exhibit a strong linear correlation, thus allowing for efficient coupling. A crossed four-bar mechanism \cite{yoon2017underactuated} is employed by interconnecting multiple segments to achieve coordinated motion between these joints. To avoid undesired coupling between the PIP and DIP joint movements and to maximize the available internal space within the finger, the actuator is placed inside the proximal phalanx. The motor's motion is transmitted to the connecting rod through a worm gear system, thereby achieving a compact structure that allows for self-locking of the position. The schematic diagram of the mechanism is depicted in Fig. \ref{FigureLabel3}b and \ref{FigureLabel3}f. By adding only one link, this mechanism efficiently achieves underactuation of the two joints. To replicate the natural motion trajectory of the human finger, we employed the Quantum Mocap Metagloves (MANUS) to capture the joint angles of the human hand during grasping tasks. Accordingly, the link length and position were optimized using a particle swarm optimization algorithm. Figure \ref{FigureLabel3}c illustrates the optimized fingertip motion trajectory and corresponding trajectory of the human finger (the red dashed curve), demonstrating a close similarity between the proposed mechanism and the kinematic characteristics of the human finger.

The human MCP joint is characterized by two degrees of freedom, enabling both Flex/Ext and Abd/Add movements. This biaxial design is crucial for dexterous robotic hands, as it allows the fingers to perform a wide range of grasping patterns, adapt to object geometry with enhanced stability, and execute fine in-hand manipulations that would be unattainable with only one rotational axis. Two main types of transmission schemes are widely employed to realize such 2-DOF joints in anthropomorphic robotic hands: differential-based actuation and linkage-driven actuation. In the HIT/DLR hand \cite{liu2008multisensory}, the 2-DOF MCP joint is driven by two motors through a bevel-gear differential assembly integrated at the finger base. Flex/Ext is produced by the synchronous rotation of both motors, while Abd/Add is produced by their counter-rotation. The ILDA hand \cite{kim2021integrated}, by contrast, employs a parallel-serial hybrid linkage in which two palm-mounted linear actuators drive both rotational axes: the 2-DOF MCP motion is realized through two PSS (Prismatic-Spherical-Spherical) kinematic chains, whose differential and common-mode displacements independently govern Abd/Add and Flex/Ext. Nevertheless, both schemes share inherent limitations. They require two motors to operate cooperatively, which increases control complexity. The coupled kinematics restricts independent axis control, thereby limiting flexibility in the arrangement of DOFs. Moreover, the differential-based or linkage mechanisms tend to be bulky, posing difficulties for integration into constrained space comparable in size to human hands.

Here, we propose a spatial four-bar spherical linkage mechanism aimed at effectively decoupling the bidirectional motion of the MCP joint. As illustrated in Fig. \ref{FigureLabel2}d and \ref{FigureLabel3}b , the mechanism consists of an offset bar and a tilting link, forming a compact linkage chain. This spherical four-bar arrangement offers a simple and compact solution that provides enhanced flexibility in movement configuration. The orthogonal rotational axes allow for completely independent actuation of Flex/Ext and Abd/Add movements, which is challenging to realize with conventional mechanisms. Upon actuation, the motor's rotational motion is transmitted to the offset bar, which subsequently modulates the inclination angle of the tilting link. The finger joint, mounted on the upper surface of the tilting link, thereby performs controlled flexion and extension movements. The schematic diagram and kinematic modeling of the structure is further presented in Fig. \ref{FigureLabel3}g.

Owing to the kinematic decoupling inherent in the spatial linkage mechanism, the flexion of the MCP joint can be integrated with any form of Abd/Add transmission mechanism, provided that the axes of Abd/Add and Flex/Ext intersect at the spherical center. Here, in consideration of the compact spatial constraints within the palm, a planar four-bar linkage is employed to achieve large-range lateral motion at the finger base joint. The schematic diagram of the mechanism is depicted in Fig. \ref{FigureLabel3}h. A micro servo motor (KST/X06) embedded within the palm actuates the motion of the mechanism. The motor's output is transmitted through the planar four-bar linkage mechanism connected to the base joint of each finger. By optimizing the lengths of the linkages, this design emulates natural human finger Abd/Add patterns.

\subsection{Thumb Mechanism}

The thumb, being the most dexterous and powerful digit of the human hand, plays a critical role in enabling stable and effective grips by opposing the other fingers \cite{wang2017thumb}. Due to its unique structure and position, the thumb necessitates a distinct design compared to the other fingers. The human thumb, characterized by its two exposed joints, is enabled by the first metacarpal bone to perform a variety of movements including rotation and flexion, thereby meeting requirements for diverse grasping and manipulation tasks. Therefore, to preserve the reconfigurable relationship between the thumb and the palm, a three-joint structure is implemented. The structure design and its exploded view are shown in Fig. \ref{FigureLabel2}e and \ref{FigureLabel2}f.

Regarding the degrees of freedom configuration, the thumb is designed with three Flex/Ext DOFs and one Abd/Add DOF. The Flex/Ext motors are located within the proximal phalanx and the first metacarpal, and drive a shaft system through worm gear mechanisms (see Fig. \ref{FigureLabel3}d). Analogous to the Abd/Add mechanism employed for the remaining four digits, the thumb Abd/Add is actuated by a servo motor fixed within the palm via a planar four-bar linkage. This lateral movement adjusts the relative position between the thumb and palm, with its kinematic workspace and anatomical arrangement emulating the human CMC joint. Consequently, this anatomical configuration allows the thumb to oppose and establish independent contact with each of the other four fingers, enabling various pinch grasp modalities.

\subsection{Control Strategy}

The control of high-degree-of-freedom robotic hands presents significant challenges due to the need for coordinated actuation and real-time feedback across multiple joints. To address this, a compact and cost-effective control architecture is developed, integrating heterogeneous actuators and distributed embedded control. 

In the proposed design, the hand incorporates a total of 16 actuators, consisting of 11 brushed DC motors and 5 servo motors. Each brushed motor is equipped with a Hall-effect encoder, providing real-time angular position feedback for closed-loop control. To achieve simple and cost-effective control of them, two ESP32 Dev Module microcontrollers are employed. Servo motors are driven via pulse-width modulation (PWM) signals, where the duty cycle directly determines their angular positions. In contrast, the brushed DC motors are regulated with a closed-loop position control scheme, where encoder feedback is used to estimate joint angles and compute position error relative to desired setpoints. A proportional–integral–derivative (PID) controller running on the microcontroller generates control signals, which are delivered to L9110s motor drivers to actuate the motors.

The entire control system is highly integrated within a compact volume of 52mm × 52mm × 40mm, allowing seamless embedding into the wrist module of the robotic hand. The two microcontrollers communicate with the host computer via serial port. An interactive user interface provides intuitive control over individual actuators as well as coordinated multi-joint movements. Importantly, the system supports simultaneous closed-loop control of all brushed motors, ensuring stable, precise, and synchronized manipulation across multiple degrees of freedom.

\section{Kinematics Modeling}

This section presents the kinematic modeling of the representative mechanisms implemented to realize the joint motions. The forward and inverse kinematics of the linkage-driven mechanisms are derived to establish the mapping between the actuator space and the joint space, while characterizing their kinematic and mechanical behaviors. Building on this, the joint ranges of motion, fingertip trajectories, and reachable workspaces are further analyzed to evaluate the anthropomorphic kinematic performance of the proposed hand.

\subsection{Four-Finger Coupled Joint Mechanism}

The human finger is characterized by the inherent coupled motion of its PIP and DIP joints. To replicate this movement, a crossed four-bar linkage is utilized, as schematically simplified in Fig. \ref{FigureLabel3}f. The mechanism satisfies the closed‑loop vector equations:
\begin{equation}
\begin{cases}
l_j \cos\theta_p = l_m \cos\theta_m + l_d \cos\theta_d, \\[4pt]
l_j \sin\theta_p = l_p + l_m \sin\theta_m + l_d \sin\theta_d,
\end{cases}
\end{equation}
where in this configuration, $l_{m}$ denotes the driven link that determines the motor angle of $\theta_{m}$, and $l_{j}$ is the length of the proximal phalanx, while $l_{p}$ and $l_{d}$ represent the fixed pivot distances on the proximal and distal phalanges. The angles $\theta_{p}$ and $\theta_{d}$ correspond to the rotations of the PIP and DIP joints, respectively.

Given the known lengths of $l_{m}$, $l_{j}$, $l_{p}$, and $l_{d}$, along with the range of each angle, the relationship between the motor's rotation angle and the rotational angles of the two joints can be derived. The detailed derivation process is given as follows.

Squaring and summing the two equations in Eq. (1) eliminates \(\theta_p\) and, upon introducing the constant, yields
\begin{equation}
l_ml_d\cos\theta_m\cos\theta_d + l_d(l_p+l_m\sin\theta_m)\sin\theta_d = -l_ml_p\sin\theta_m - C.
\end{equation}
where
\begin{equation}
C = \frac{l_m^2+l_p^2+l_d^2-l_j^2}{2},
\end{equation}



The left‑hand side can be written as \(l_d R\cos(\theta_d-\varphi)\) with
\begin{equation}
\begin{aligned}
R &= \sqrt{l_p^2 + l_m^2 + 2 l_m l_p \sin\theta_m}, \\[4pt]
\varphi &= \operatorname{atan2}(\; l_p+l_m\sin\theta_m,l_m\cos\theta_m).
\end{aligned}
\end{equation}

Thus, Eq. (1) becomes
\begin{equation}  
l_d R \cos(\theta_d-\varphi) = -l_ml_p\sin\theta_m - C.
\end{equation}


Solving for \(\theta_d\) and restricting to the value within the prescribed angle ranges, the explicit expression is given by
\begin{equation}
\theta_d = \varphi \pm \arccos\!\left(\frac{-l_ml_p\sin\theta_m - C}{l_d\sqrt{l_p^2+l_m^2+2l_ml_p\sin\theta_m}}\right).
\end{equation}
Once \(\theta_d\) is known, \(\theta_p\) follows directly from the original equations:
\begin{equation}
\theta_p = \operatorname{atan2}\!\left(l_p + l_m\sin\theta_m + l_d\sin\theta_d,\; l_m\cos\theta_m + l_d\cos\theta_d\right).
\end{equation}

The \(\pm\) sign in the expression for \(\theta_d\) corresponds to two possible assembly modes. The physically admissible solution must satisfy the prescribed ranges for \(\theta_m,\theta_p,\theta_d\). The relationship among the three angles can thus be computed.

\subsection{Spatial Spherical Mechanism}

The spherical four-bar mechanism features all revolute joint axes intersecting at a common point in space (the spherical center), constraining all points of the mechanism to move on concentric spherical surfaces centered at that point. This inherent geometry enables motion decoupling. When actuated to perform bending motions, the mechanism induces a change solely in the bending angle, with negligible effect on the Abd/Add axis, and vice versa. This decoupling characteristic significantly simplifies the control algorithm, enabling the end-effector to adjust its orientation independently and precisely along two axes. A schematic diagram of the spatial linkage mechanism is depicted in Fig. \ref{FigureLabel3}g.

We define the two endpoints of the flexion link as $A$ and $B$, with the common center point $O$. The vectors $\boldsymbol {OA}$ and $\boldsymbol {OB}$ maintain a fixed spatial angle $\alpha$ between them. By design, when the motor angle is zero, the finger is oriented vertically. In this configuration, as the motor rotates to an angle $\psi$, the coordinates of $\boldsymbol{OA}$ can be expressed as:
\begin{equation}
\mathbf{OA} = \left( -l_a \cos\alpha \sin\psi,\; -l_a \cos\alpha \cos\psi,\; -l_a \sin\alpha \right),
\end{equation}
where $l_{a}$ denotes the length of vector \(\boldsymbol{OA}\). 

When the proximal phalanx bends by an angle \(\theta_f\) and deviates laterally by \(\phi_s\), the coordinates of point \(B\) can be expressed based on geometric relationships as:
\begin{equation}
\mathbf{OB} = \left( l_b \sin\phi_s ,\; -l_b \cos\theta_f \cos\phi_s,\; l_b \sin\theta_f \cos\phi_s \right),
\end{equation}
where $l_{b}$ denotes the length of vector \(\boldsymbol{OB}\). 

Since the angle \(\alpha\) between the links remains constant, applying the law of cosines gives:

\begin{align}
\cos\alpha &= \frac{\boldsymbol{OA} \cdot \boldsymbol{OB}}{|\boldsymbol{OA}|\,|\boldsymbol{OB}|} = \cos\alpha \cos\psi \cos\theta_f \cos\phi_s \nonumber \\
&\qquad -\cos\alpha \sin\psi \sin\phi_s  - \sin\alpha \sin\theta_f \cos \phi_s,
\label{eq:angle_relation}
\end{align}

which leads to the following constraint equation:
\begin{equation}
\cos\phi_s \cos\psi \cos\theta_f
- \tan\alpha \cos\phi_s \sin\theta_f
= 1 + \sin\phi_s \sin\psi
\end{equation}

The motor rotates within the range $[-\pi,\pi]$. Therefore, for a given motor rotation angle $\psi$ and measured Abd/Add angle $\phi_{s}$, the joint flexion angle can be obtained via forward kinematics as:
\begin{equation}
\begin{split}
\theta_f = \pm \arccos\left(\frac{1+\sin\phi_s\sin\psi}{\sqrt{\cos^2\phi_s\cos^2\psi+\tan^2\alpha\cos^2\phi_s}}\right) \\ + \arctan\left(\frac{-\tan\alpha}{\cos\psi}\right),\quad \theta_f \in [0,\ \frac{\pi}{2}].
\end{split}
\end{equation}

For given target bending angle $\theta_{f}$ and lateral deflection angle $\phi_{s}$, the required motor angle $\psi$ can be derived as:
\begin{equation}
\begin{split}
\psi =  \pm \arccos\!\left(\frac{\cos\alpha + \sin\alpha \sin\theta_f \cos\phi_s}{\cos\alpha \sqrt{\cos^2\theta_f \cos^2\phi_s + \sin^2\phi_s}}\right) \\
       + \operatorname{arctan2}\!\left(-\sin\phi_s,\ \cos\theta_f \cos\phi_s\right),\quad \psi \in [-\pi,\ \pi].
\end{split}
\end{equation}

 Kinematic analysis demonstrates that the flexion angle of the finger joint varies proportionally with the motor's rotational input, achieving a maximum flexion of approximately 60°. As illustrated in Fig. \ref{FigureLabel3}g, the relationships among the motor rotation angle, Abd/Add angle, and Flex/Ext angle are presented. The results show that, for a given motor rotation angle, variations in the lateral angle have only a limited influence on the joint flexion position, indicating effective motion decoupling between the two movements. Furthermore, the mechanism exhibits nonlinear characteristics in both motion transmission and torque generation. Specifically, the output torque decreases initially and subsequently increases throughout the motion cycle. Hence, during the initial movement phase, the output torque remains relatively low, facilitating rapid and energy-efficient positioning. In contrast, during the grasping phase, the torque increases substantially to enable secure object holding. This nonlinear torque characteristic is particularly advantageous for power grasping tasks, as it naturally aligns with the functional requirements: low torque is sufficient for finger motion during the approach, while high torque is essential for maintaining stable grip against external disturbances. The spherical four-bar mechanism thus offers inherent mechanical advantage, compactness, and motion decoupling, making it well-suited for anthropomorphic robotic fingers where both dexterity and load-bearing capacity are required.

\subsection{Lateral Motion Mechanism}

The Abd/Add motions of both the thumb and the other four fingers are actuated by servo motors through planar four-bar linkages. The kinematic model of these linkages is analogous to that of the crossed four-bar mechanism. A schematic representation of the mechanism is illustrated in Fig. \ref{FigureLabel3}h. Taking the middle finger as an instance. The relationship between the servo motor angle $\theta_{s}$ and the resulting rotational angle of the middle finger $\theta_{b}$ satisfies the following vector equations:
\begin{equation}
\begin{cases}
x_b = l_s \cos\theta_s + l_c \cos\theta_c + l_b \cos\theta_b, \\[4pt]
y_b = l_s \sin\theta_s + l_c \sin\theta_c + l_b \sin\theta_b,
\end{cases}
\end{equation}
where $l_s$, $l_c$, and $l_b$ denote the lengths of the servo arm, the connecting link, and the base joint link, respectively. $\theta_c$ is an intermediate variable introduced to simplify the calculation. $(x_b, y_b)$ denotes the coordinates of the base joint lateral axis.

Following the similar calculation, the relationship between the finger lateral angle $\theta_{b}$ and the servo angle $\theta_{s}$ can be expressed as:
\begin{equation}
l_b R \cos(\theta_b-\varphi) = C - x_bl_scos\theta_s-y_bl_ssin\theta_s .
\end{equation}
where,
\begin{equation}
\begin{aligned}
C &= \frac{x_b^2 + y_b^2 + l_s^2 + l_b^2 - l_c^2}{2}, \\[4pt]
R &= \sqrt{x_b^2 + y_b^2 + l_s^2
- 2l_s \bigl(x_b \cos\theta_s + y_b \sin\theta_s\bigr)}, \\[4pt]
\varphi &= \operatorname{atan2}
(y_b-l_s\sin\theta_s,\; x_b-l_s\cos\theta_s).
\end{aligned}
\end{equation}

With this equation, the relationship between the Abd/Add angles of all fingers and the servo rotational angles, and their range of motion can be obtained. The results indicate that the thumb can achieve a lateral movement range of $90^\circ$, while the other fingers all exhibit more than $30^\circ$ movement ranges. Compared with the human hand, the proposed robotic hand effectively reproduces a comparable workspace.

\subsection{Workspace}

The reachable workspace of a robotic hand refers to the complete set of spatial positions that the end-effector can access through all feasible joint configurations. It serves as a quantitative metric describing the operational boundaries and spatial capabilities of the hand, defining the physical limits within which manipulation tasks can be executed. Based on the joint motion ranges derived from forward kinematics and the overall configuration of the hand, the Denavit–Hartenberg (DH) parameters for each finger and the entire hand can be established, thereby enabling the determination of their respective workspaces. The results are shown in Fig. \ref{FigureLabel3}c, \ref{FigureLabel3}e, and \ref{FigureLabel3}i. The results indicate that the implemented mechanism endows both the fingers and the thumb with human-like motion trajectories and workspaces. Moreover, the thumb exhibits the highest flexibility, as its workspace intersects with those of the other four fingers, thereby enabling effective opposition movements. The hand’s workspace adequately covers the regions required for typical grasping and manipulation tasks.

\begin{table}[t]
\centering
\caption{Comparison of functional range of motion (RoM) between the proposed hand and the human hand\cite{barakat2013range, bain2015functional}.}
\label{tab:rom_comparison}
\renewcommand{\arraystretch}{1.2}
\begin{tabular}{cccc}
\hline
\textbf{Part} & \textbf{Joint/Category} & \textbf{Our Hand ($^\circ$)} & \textbf{Human Hand ($^\circ$)} \\
\hline

\multirow{4}{*}{Thumb}
& CMC Abd/Add           & 90 & 73.1 \\
& CMC Flex/Ext          & 95 & 61.2 \\
& MCP Flex/Ext          & 100  & 68.1 \\
& IP Flex/Ext           & 80  & 100.0 \\
& \textbf{Total RoM} & \textbf{365} & \textbf{302.4} \\
\hline

\multirow{5}{*}{Four fingers}
& MCP Abd/Add           & 30 - 40  & 25 - 30 \\
& MCP Flex/Ext          & 60 & 71.0 \\
& PIP Flex/Ext          & 90  & 87.0 \\
& DIP Flex/Ext          & 50  & 64.0 \\
& \textbf{Total RoM} & \textbf{230 - 240} & \textbf{247 - 252} \\
\hline

\end{tabular}
\end{table}

\section{Performance Evaluation}

\begin{figure}[!t]
    \centering
    \includegraphics[width=1\linewidth]{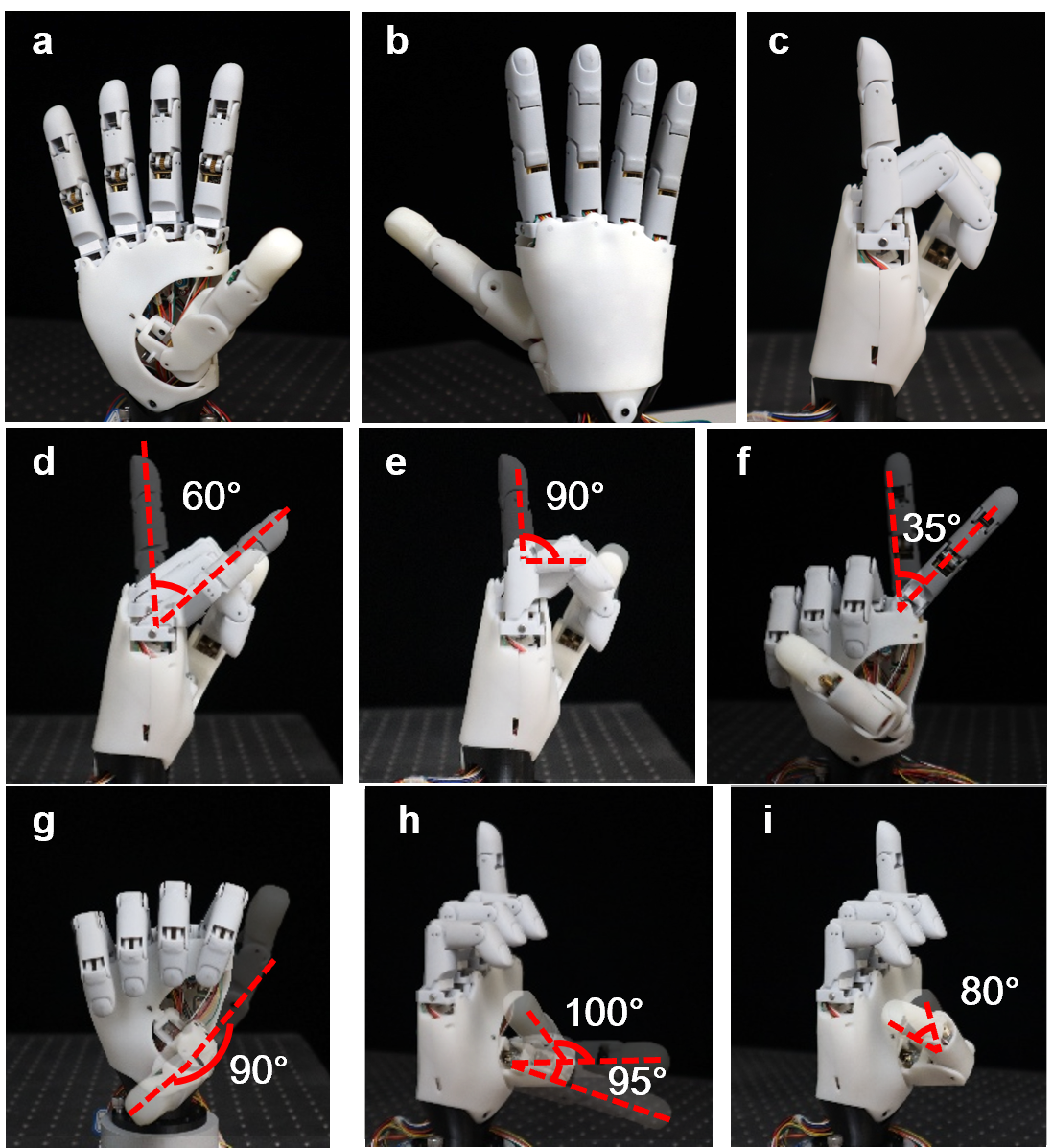}
    \caption{The manufactured hand and the range of motion of its individual joints. (a) Front view; (b) Rear view; (c) Side view; (d-i) Maximum range of motion for each joint.}
    \label{FigureLabel4}
\end{figure}

\begin{figure}[ht]
    \centering
    \includegraphics[width=1\linewidth]{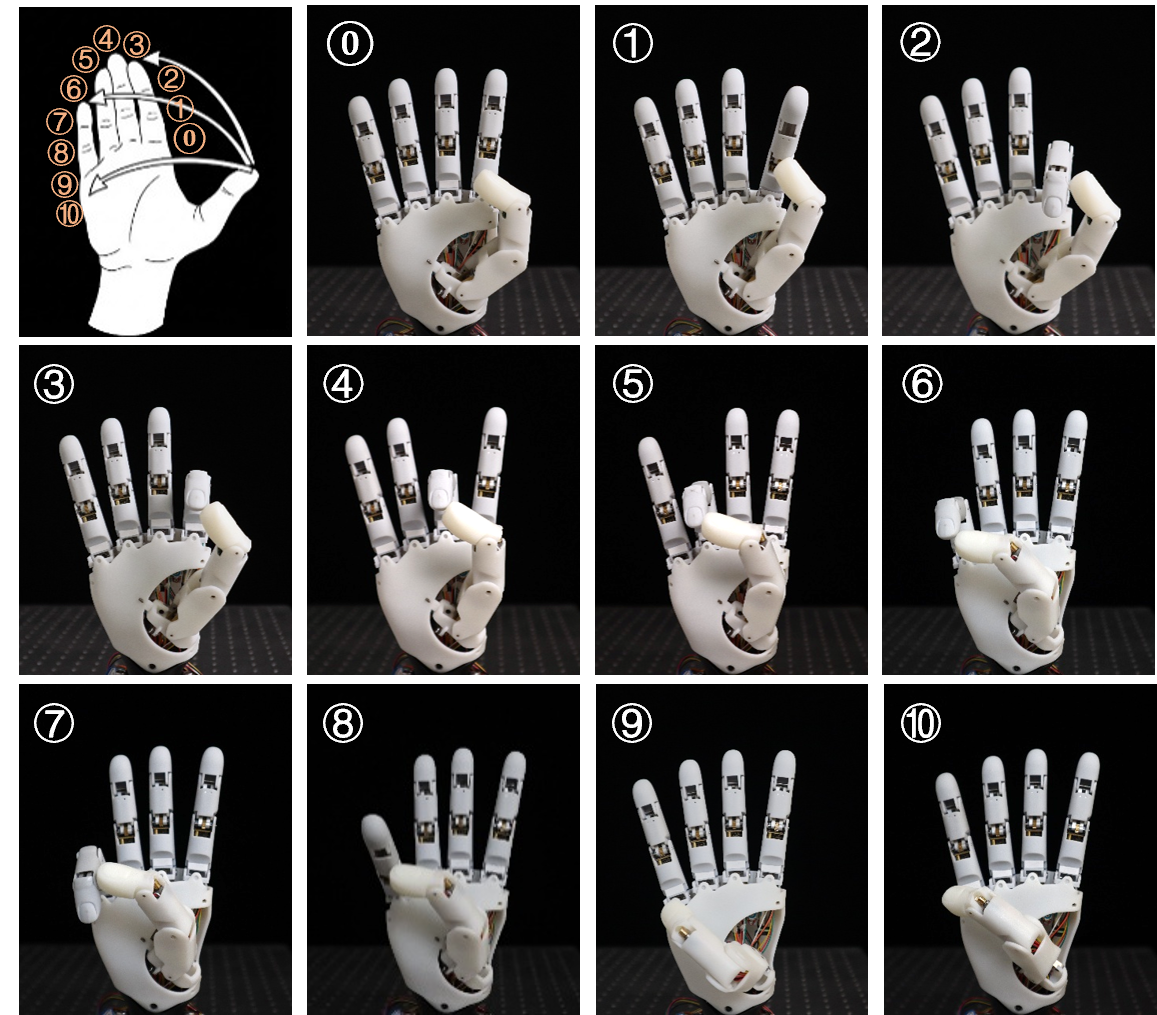}
    \caption{The hand achieves the highest possible score in the Kapandji test owing to its dexterous, opposable thumb which is able to reach all ten locations on the hand.}
    \label{FigureLabel5}
\end{figure}

To systematically evaluate the proposed proof-of-concept prototype, a series of experimental tests was conducted. The primary objective is to validate the hand's core design goals of anthropomorphic dexterity and functional versatility. In this section, we evaluate the performance of this hand from four aspects: movement dexterity, force and load characteristics, grasping performance, and tool manipulation capability.

\subsection{Finger Dexterity and Thumb Opposability}

The prototype of the dexterous hand is presented in Fig. \ref{FigureLabel4}a-c, while the motions and corresponding working ranges of the five fingers are illustrated in Fig. \ref{FigureLabel4}d–i. 
These experimentally measured joint motions are consistent with the kinematic analysis results. In addition, the degrees of freedom of the hand can be actuated independently or coordinately, enabling versatile and anthropomorphic manipulation behaviors. To further evaluate the motion capability of the proposed hand, the RoM of each joint was benchmarked against the reported RoM of the human hand, as summarized in Table III. The comparison results indicate that the proposed dexterous hand is capable of reproducing most major human finger motions within a comparable range. 

Here we use the Kapandji test \cite{KAPANDJI198667} to evaluate the opposition performance of the thumb. This standardized assessment, originally developed for human hand function, was employed to validate the thumb's capacity to touch the tips of the other fingers and the palmar region across a graded scale. As illustrated in Fig. \ref{FigureLabel5}, the proposed hand can perform the full range of opposition gestures, from simple fingertip contact to complex pulp-to-pulp grips. Moreover, the resulting thumb-finger confrontation postures closely mimic those of the natural human hand. The results confirm that the kinematic structure and actuation strategy of the thumb enable a level of dexterity comparable to the key opposition capabilities of the human hand, thereby covering a substantial portion of the grasping and manipulation range observed in humans.

\subsection{Force and Load Performance}

\begin{figure}[!t]
    \centering
    \includegraphics[width=1\linewidth]{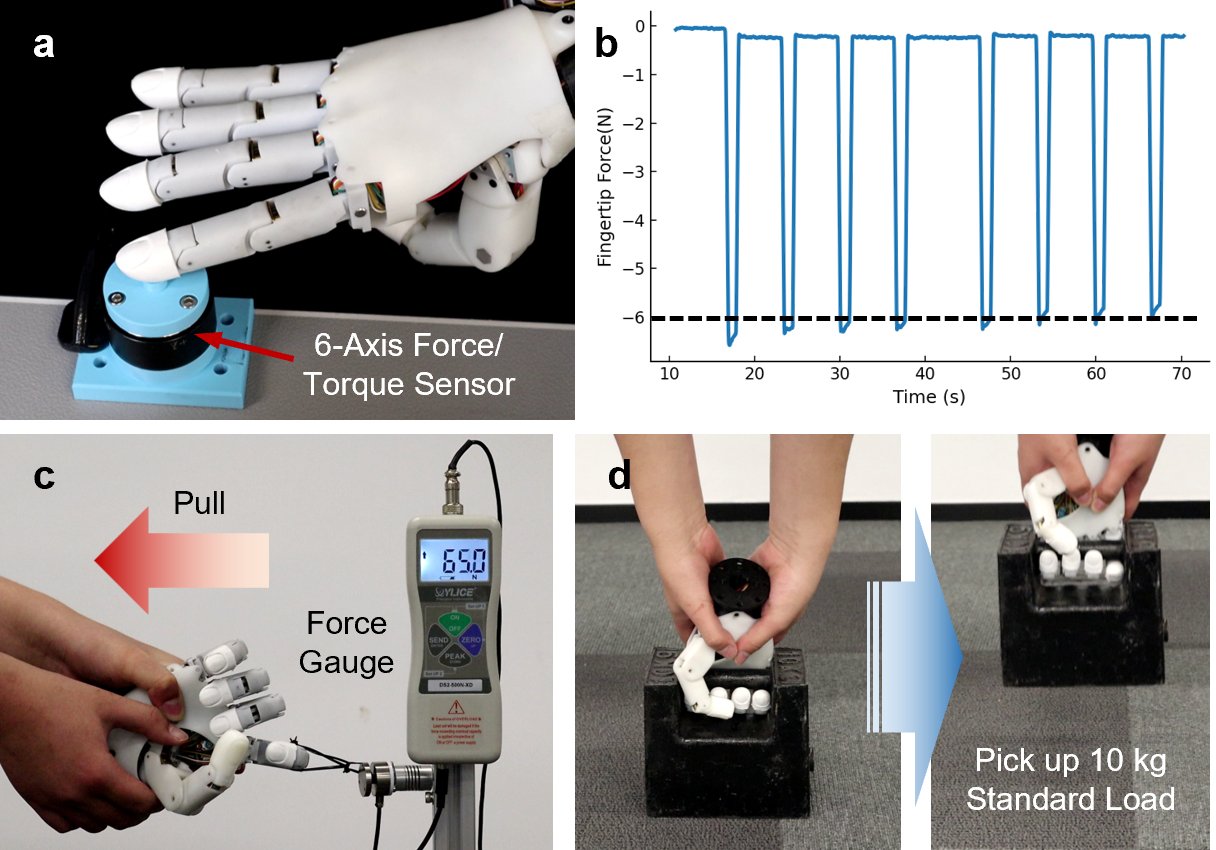}
    \caption{Active output force and payload capacity test of the hand. (a) The experimental setup for measuring active output force; (b) The results of repeated active fingertip force tests; (c) The experimental platform for testing ultimate load, with a single finger capable of withstanding forces exceeding 65N; (d) The hand passively holds a 10kg dumbbell.}
    \label{FigureLabel6}
\end{figure}

Here, we evaluated the force and load performance of the hand from two aspects: fingertip active force, and passive payload capacity. First, the robotic hand was mounted on a UR5e robotic arm, with its index finger actuated to apply a contact force against a commercial force sensor ($\gamma$45, DAYSENSOR). The experimental setup is illustrated in Fig. \ref{FigureLabel6}a Two motors were commanded to flex the MCP and PIP joints of the finger under a rated voltage of 12 V, hold the flexed position for approximately 1 s, and then return to the initial position. This procedure was repeated eight times. As shown in Fig. \ref{FigureLabel6}b, the fingertip consistently exerted a peak force exceeding 6N across all cycles, demonstrating stable and repeatable fingertip actuation.

All Flex/Ext motions of the finger joints are driven by linkage or worm-gear mechanisms with inherent self-locking characteristics. Thus, it is necessary to evaluate their passive load-bearing capacity. Accordingly, we conducted tests on both the load strength of a single finger and the lifting capacity of the whole hand. As shown in Fig. \ref{FigureLabel6}c, one end of the cable tie was fastened to the middle phalanx, with the other end attached to a digital force sensor; a pulling force was then applied. The results showed that the finger could maintain its position under a load of approximately 65N. Fig. \ref{FigureLabel6}d further demonstrated that the hand can support and lift the 10kg mass. Under this condition, no noticeable finger opening, slippage, or structural instability was observed, indicating strong static load-bearing capability.

\subsection{Grasp Capacity Performance}

\begin{figure*}[!htbp]
   \centering
   \includegraphics[width=1\linewidth]{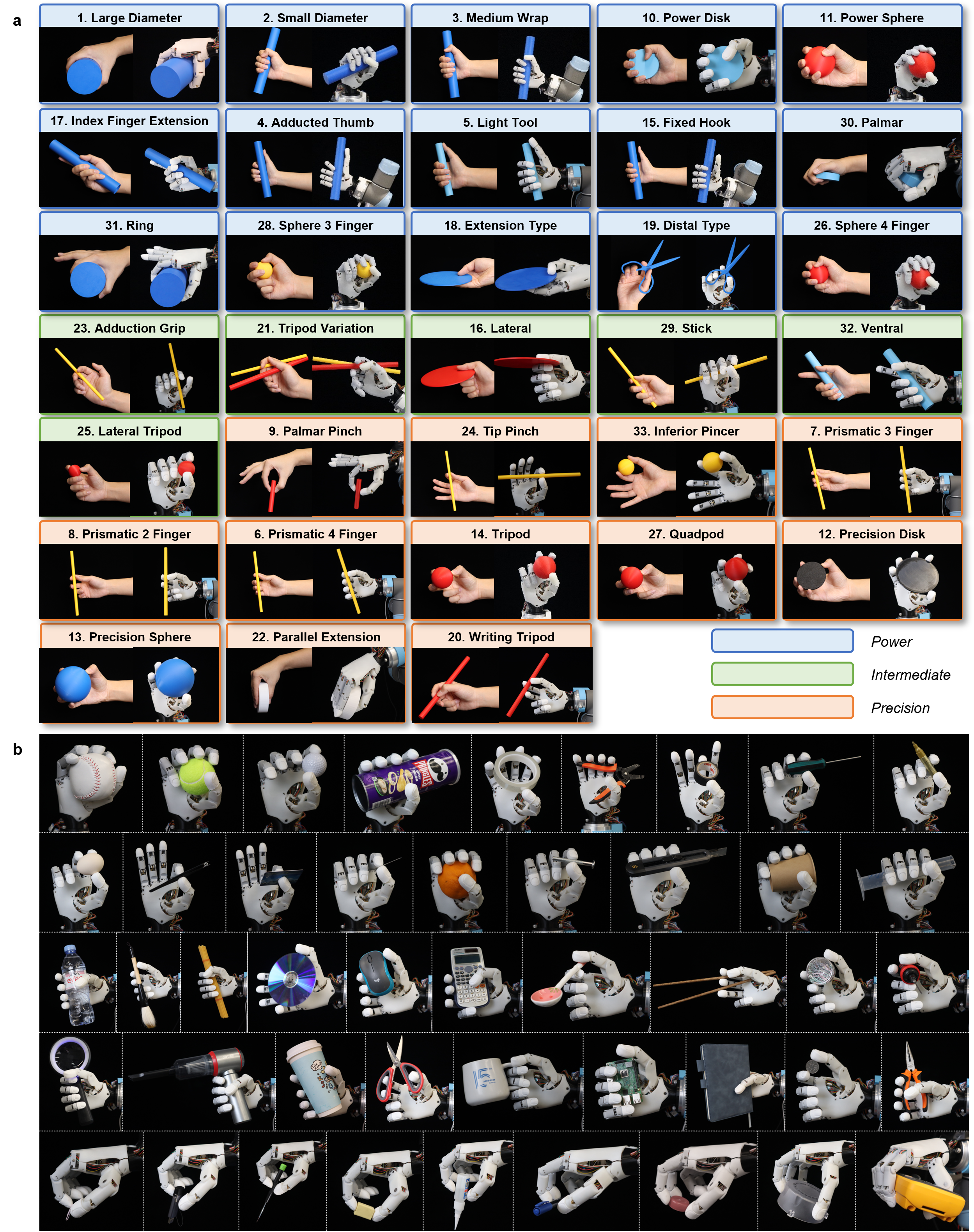}
    \caption{Versatile grasping modalities and generalized prehensile performance of the hand. (a) The prototype is able to replicate all 33 grasp postures of the most comprehensive used GRASP taxonomy. The blue, green, and orange backgrounds represent the power, intermediate, and precision grasp categories, respectively; (b) Grasping capability of the hand when handling everyday objects in diverse postures and orientations.}
    \label{FigureLabel7}
\end{figure*}

The grasping capability of the robotic hand, particularly in handling objects of varying shapes, weights, sizes, and thicknesses, serves as a crucial metric for evaluating its functionality. The grasping taxonomy proposed by Feix et al. \cite{feix2015grasp} was adopted here to evaluate the dexterity and functionality of the hand. This taxonomy consists of 33 typical grasping patterns that can cover the majority of grasp types frequently used in daily life. All grasp objects (sphere, cylinder, disc, scissors) were fabricated using 3D printing. As depicted in Fig. \ref{FigureLabel7}a, the hand can successfully perform all 33 grasp types and achieve stable grasping across varying poses and orientations.

To further demonstrate the hand's grasping capability, we selected a representative set of everyday objects as grasping targets. The hand was commanded to perform corresponding grasp postures, mimicking the way human hands handle these objects. The results are shown in Fig. \ref{FigureLabel7}b. With its anthropomorphic structural design, the hand consistently achieved stable grasps across the 46 selected daily objects while also effectively replicating the functional versatility characteristic of human manual dexterity. These results highlight the hand's remarkable adaptability to objects with diverse characteristics, as well as its potential for practical applications when appropriate grasping strategies are employed.

\begin{figure*}[!htbp]
    \centering
    \includegraphics[width=1\linewidth]{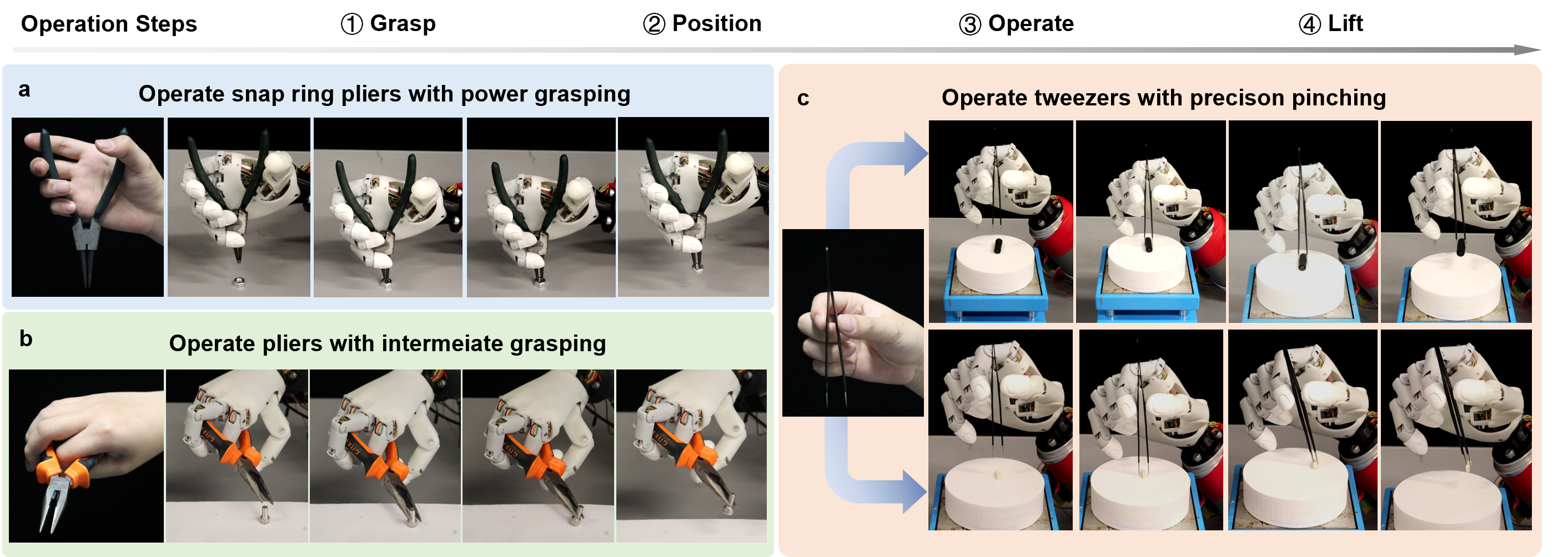}
    \caption{Demonstration of dexterous hand manipulation capabilities through tool operations. (a) Clamp a nut with snap ring pliers; (b) Grip a screw with pliers; (c) Pick up the pen cap and the connector with tweezers.}
    \label{FigureLabel8}
\end{figure*}

\subsection{Task-Oriented Tool Operation}

Achieving the capability to manipulate everyday tools for specific tasks has long been a key objective in dexterous hand research. Unlike grasping, which primarily concerns the secure gripping and holding of objects; manipulation, in contrast, requires precise and coordinated multi-joint control to achieve task-oriented motions. Meeting this demand presents significant challenges, as it necessitates real-time feedback, adaptive regulation, and fine sensorimotor coordination. To validate the dexterity of the proposed hand, three simple tool manipulation demonstrations were conducted. With the prototype mounted on a UR5e robotic arm, these experiments assessed the hand's ability to handle tools that require distinct grasping strategies, multi-finger coordination, and precision control. For each of the three manipulations, we manually decomposed the action sequence of operating the tool into several discrete motion positions. By controlling all joints of the dexterous hand to move simultaneously to these target positions, we were able to simulate a continuous motion, thereby achieving both holding and operation of the tool.

As illustrated in Fig. \ref{FigureLabel8}a, the first experiment involves using snap ring pliers to clamp a nut. The hand adopts a power‑grasp posture, with the thumb pressing against one handle and the other four fingers against the opposite handle. Multi‑finger coordinated flexion generates the grip force needed to open the pliers' jaw, allowing the nut to be securely lifted from the table. This task showcases the hand's effective multi‑finger coordination and sufficient power grip strength. 
In the plier operation experiment, the hand secures one end of the pliers with the index finger and thumb, while the other end is held by alternating the index and middle fingers in an interlocking manner. The opening and closing of the pliers' jaws are achieved through coordinated flexion and extension of the index and middle fingers. As shown in Fig. \ref{FigureLabel8}b, this configuration enables the hand to pick up a screw from the table and subsequently release it, highlighting the hand's capacity for fine, finger‑synchronized tool handling. The third demonstration illustrates the hand's ability to perform fine manipulation using tweezers to grasp a tiny component. This task requires precise pinching between the index finger and thumb, as the tweezers demand delicate compression force and accurate alignment to securely capture the small object. Fig. \ref{FigureLabel8}c shows the process of picking up a pen cap and a connector, highlighting the hand’s precise manipulation capability.

Collectively, the results illustrate the proposed hand's versatility in handling tools with varying mechanical requirements, ranging from force‑driven actuation to precision gripping. The results confirm that, when combined with appropriate control, the hand is capable of supporting tool-use behaviors and executing task‑oriented manipulations that extend beyond simple grasping. This capability advances its suitability for practical dexterous applications in human-centered scenarios.

\section{Conclusion}

In this work, we present a compact, low-cost, linkage-driven anthropomorphic hand that achieves human-like dexterity and functional versatility through a mechanically efficient and biomimetic design. The proposed design demonstrates that the long-standing trade-off among dexterity, compactness, and affordability in robotic hand design can be addressed through principled mechanism design and embedded mechanical intelligence. To achieve this objective, the proposed hand integrates 20 physical joints with 16 actively actuated DOFs within a human-hand-scale form factor while maintaining a lightweight structure of 320g and a hardware cost below USD 400. By combining spatial spherical linkage mechanisms, planar linkage transmissions, and worm-gear self-locking structures, the system realizes decoupled multidirectional motion, efficient transmission of coordinated joint behaviors, and high passive load-bearing capability without relying on bulky architectures. 

Experimental results validate the effectiveness of the proposed design from multiple aspects. The hand achieves the full Kapandji thumb opposition score, successfully reproduces all 33 grasp types defined in the Feix taxonomy, and demonstrates representative task-oriented tool manipulation capabilities requiring coordinated multi-finger control. These results confirm that compact hand-scale integration with high manipulation performance can be achieved through coordinated motion generation and embedded mechanical intelligence, while demonstrating robustness and accessibility. More importantly, the proposed system provides a scalable and cost-effective intermediate solution between simple low-DOF grippers and expensive fully actuated commercial dexterous hands, making high-dexterity robotic manipulation more accessible for robot learning, teleoperation, and human-centered robotic applications.

Future work will focus on integrating multimodal tactile sensing and teleoperation interfaces to further advance the hand’s perceptual and operational autonomy. First, real-time acquisition of human joint angles using the data glove will be implemented to enable intuitive teleoperation, allowing the hand to replicate natural human hand motions with high fidelity. Second, tactile sensing arrays will be integrated onto the fingertip surfaces and the palm to capture rich contact information. Third, structural optimizations will be pursued to further expand the reachable workspace and improve kinematic performance. These enhancements will transform the hand into a sensorized, teleoperable platform for real-world robot learning and human-centered dexterous manipulation.




\bibliographystyle{IEEEtran}
\bibliography{reference.bib}

\end{document}